\documentclass[runningheads]{llncs}
\usepackage[T1]{fontenc}

\usepackage{authblk}

\usepackage{lineno,hyperref}
\usepackage{color,array}
\usepackage{amsmath,amssymb}
\usepackage{subfigure}
\usepackage{threeparttable}
\usepackage{multirow}

\usepackage{graphicx}

\begin{document}
\title{Training-free Clothing Region of Interest Self-correction for Virtual Try-On}
%
%
\author{
Shengjie Lu\inst{1}\orcidID{0009-0002-8922-2488} \and
Zhibin Wan \thanks{These authors contributed equally to this work}\inst{1}  \and
Jiejie Liu\inst{2} \and
Quan Zhang\inst{2} \and
{Mingjie Sun \thanks{corresponding author}}\inst{1}
}

%
%
\institute{Department of Computer Science and Technology, Soochow University, Suzhou, China \and
School of Advanced Technology, Xian Jiaotong-Liverpool University, Suzhou, China
\\
\email{20235227111@stu.suda.edu.cn}
\email{20234227071@stu.suda.edu.cn}
\email{jiejie.liu22@student.xjtlu.edu.cn}
\email{quan.zhang@xjtlu.edu.cn}
\email{mjsun@suda.edu.cn}
}

\maketitle              
\begin{abstract}
VTON (Virtual Try-ON) aims at synthesizing the target clothing on a certain person, preserving the details of the target clothing while keeping the rest of the person unchanged. Existing methods suffer from the discrepancies between the generated clothing results and the target ones, in terms of the patterns, textures and boundaries. Therefore, we propose to use an energy function to impose constraints on the attention map extracted through the generation process. Thus, at each generation step, the attention can be more focused on the clothing region of interest, thereby influencing the generation results to be more consistent with the target clothing details. Furthermore, to address the limitation that existing evaluation metrics concentrate solely on image realism and overlook the alignment with target elements, we design a new metric, Virtual Try-on Inception Distance (VTID), to bridge this gap and ensure a more comprehensive assessment. On the VITON-HD and DressCode datasets, our approach has outperformed the previous state-of-the-art (SOTA) methods by 1.4\%, 2.3\%, 12.3\%, and 5.8\% in the traditional metrics of LPIPS, FID, KID, and the new VTID metrics, respectively. Additionally, by applying the generated data to downstream Clothing-Change Re-identification (CC-Reid) methods, we have achieved performance improvements of 2.5\%, 1.1\%, and 1.6\% on the LTCC, PRCC, VC-Clothes datasets in the metrics of Rank-1. The code of our method is public at \url{https://github.com/MrWhiteSmall/CSC-VTON.git}.

\end{abstract}
\begin{figure}[!htb]
    \centering
    \includegraphics[width=1.0\linewidth]{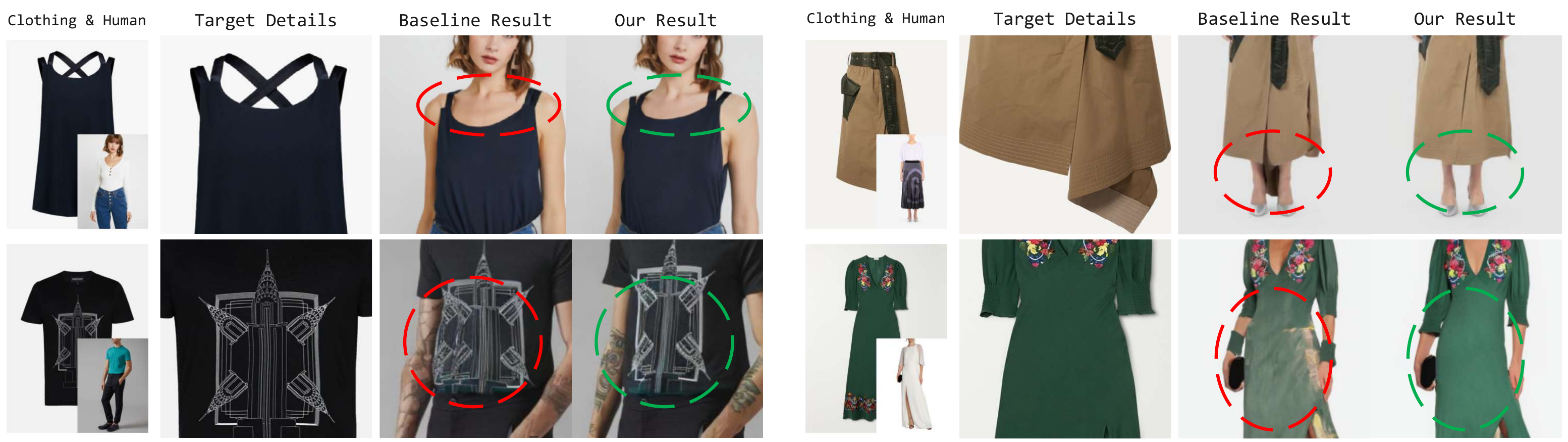}
    \caption{
        The comparison illustrates the superior accuracy of our approach when juxtaposed with the conventional baseline method. Red circles pinpoint the baseline method's inaccuracies, while green circles show our method's corrections in clothing patterns, textures, boundaries, and alignment with the person's pose, respectively, which demonstrates a significant improvement in realism and precision.}
    \label{fig:comparison}
\end{figure}
\section{Introduction}
\label{sec:intro}
The task of VTON (Virtual Try-ON) is to synthesize images of a person wearing specified clothing. The input consists of images of the target clothes and person; the output is an image of the person wearing the target clothing. This task requires clothing consistency and personal pose invariance, meaning the synthetic clothing must fit the person's body and retain cloth details such as texture, patterns, logos, \emph{etc.}.

Recent diffusion-based VTON models \cite{qiu2025noise-attach2} often use two UNets: one for extracting clothing features, named the clothing diffusion model~\cite{24-kim2024stableviton,xu2024ootdiffusion}; and another one~\cite{32-morelli2023ladi} for the final image synthesis, named the person diffusion model. The features of the clothing serve as the condition to guide the synthesis of the person diffusion model, aiming to achieve clothing consistency and personal pose invariance.
These methods often focus on how to extract better clothing features, neglecting the attention to the Clothing Region of Interest (C-RoI), \emph{i.e.}, the region in the person where the target clothes will be synthesized. This oversight makes it difficult for the introduced clothing features to achieve consistency within C-RoI~\cite{consis-1,consis-2}. We believe this is because the diffusion model, when guided by the condition, finds it challenging to restrict the synthesis operations within C-RoI. Consequently, the synthesized images exhibit the loss of texture, patterns, location and boundary details, as shown in Fig \ref{fig:comparison}.

We found that some attention maps, generated during the condition-guided denoising process, correspond to C-RoI almost in pixel-level. Thus, we attempted to restrict the attention map to concentrate on C-RoI by self-correction, enabling the model to focus on the target earlier. It allows the model to better retain the aforementioned series of features. Moreover, the proposed method focuses on each step of the denoising process, so it can be used not only during training but also as a plug-and-play module in inference.

Furthermore, traditional evaluation metrics (\emph{e.g.}, FID, KID) focus more on image generation realism, neglecting VTON's task objectives, including clothing consistency and personal pose invariance. Thus, we designed a new evaluation metric Virtual Try-on Inception Distance (VTID) to unify the evaluation of paired and unpaired datasets. 

Ultimately, our method is used to expand training data of downstream tasks like Cloth-Changing Person Re-identification (CC-ReID) \cite{ccreid-scnetguo2023}, which identifies individuals across cameras regardless of clothing changes. Our method improves the downstream model by diversifying the clothing data, reducing reliance on attire for identity recognition. Our contributions are summarized as follows:
\begin{itemize}
    \item We propose the attention-based clothing region of interest self-correction for VTON to better align with target clothing details. This modular approach is designed for easy integration without the need for additional training.
    
    \item We design a new metric VTID consistent with the evaluation objectives of the VTON task (\emph{e.g.}, clothing consistency, person pose invariance) and validate the reasonableness of this metric in subsequent test experiments.
    \item Our method surpasses previous SOTA methods with improvements of 1.4\%, 2.3\%, 12.3\%, and 5.8\% in LPIPS, FID, KID, and the new metrics VTID. Leveraging our generated data to improve the training samples of CC-Reid task in 2.5\%, 1.1\%, and 1.6\% enhancement of Rank-1 on LTCC, PRCC, and VC-Clothes datasets.
\end{itemize}

\begin{figure*}[!t]
    \centering
    \includegraphics[width=1\textwidth]{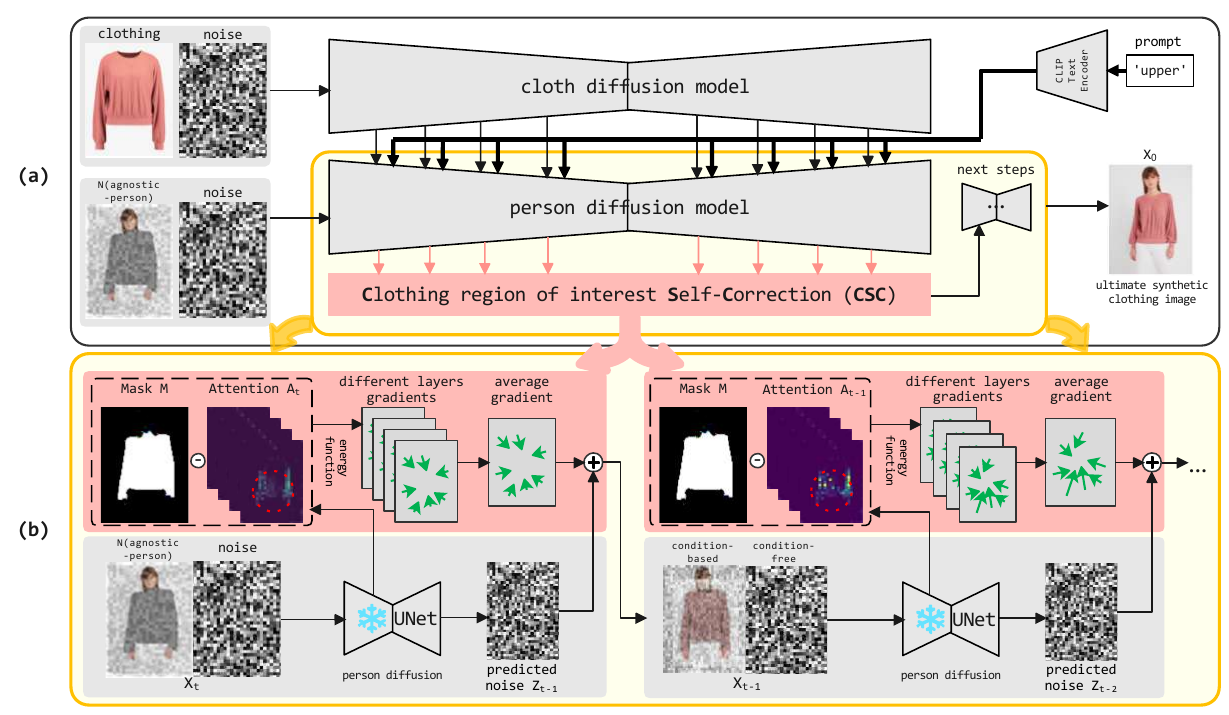}

    \caption{Illustration of our framework. (a) Apart from random noise, the clothing image/noised agnostic person is passed into the clothing/human diffusion model as the input, respectively. The clothing diffusion model transforms the clothing into a feature to guide the denoising process of the human diffusion model. (b) The proposed CSC first extracts the attention corresponding to the prompt ``upper''. Then, the difference between the attention and the mask of clothing region of interest is calculated to obtain the gradient after back-propagation. Finally, the gradient is fused with the predicted noise as correction. The corrected result serves as the input for the next step.}
    \label{fig:overview}
\end{figure*}

\section{Related Work}
\label{sec:formatting}

\subsection{Diffusion-based VTON}
%
In the early stages of VTON (Virtual Try-On Network) tasks, GANs (Generative Adversarial Networks) models \cite{6-choi2021viton,16-han2018viton,27-lee2022high} were commonly used. With the advent of Diffusion models \cite{32-morelli2023ladi,52-xie2023gp,24-kim2024stableviton}, their superiority in high-resolution generation and condition-guided generation was evident enough to challenge the era of GANs.
VTON tasks inherently require conditional control during generation. Methods vary in their conditioning approach: LaDI-VTON \cite{32-morelli2023ladi} converts clothing features to CLIP \cite{10-clip-gal2022image} embeddings via text inversion, while OOTDiffusion \cite{xu2024ootdiffusion} directly uses CLIP-encoded clothing features. StableVITON \cite{24-kim2024stableviton} instead employs cross-attention for clothing feature extraction without CLIP.

\subsection{Diffusion-based Optimization}

Optimization methods can be divided into those that require training and those that do not. For methods \cite{f31-mou2024t2i,f36-rombach2022high} that require training, Diffusion models need to address the difference between the samples used during training and those used during inference. Currently, conditions are often used to bridge this gap. For example, using text-mapped embeddings \cite{24-kim2024stableviton} as additional data allows the model to correct the results based on this control information. 
Non-training optimization methods \cite{f2-bansal2023universal,f14-hertz2022prompt,f28-meng2021sdedit,f33-parmar2023zero,f47-wang2022zero} enhance sampling by refining predicted noise at each step to better align with inputs. Current research focuses on determining optimal correction data. \cite{f14-hertz2022prompt} employs cross-attention to integrate external data with intermediate features, while DDNM \cite{f47-wang2022zero} modifies latent representations through zero-space decomposition, both operating without training.

\section{Method}

\subsection{Preliminary}

Stable Diffusion itself has been developed based on the Latent Diffusion Model (LDM) \cite{6-choi2021viton,14-gou2023taming,27-lee2022high}. During training, noise is added to $z_0$, as follows:
\begin{equation}
    q(z_t | z_0) = \mathcal{N} \big{(}z_t;
    \sqrt{\overline{\alpha}_t}z_0,
    (1-\overline{\alpha}_t)\mathbf{I}
    \big{)},
    \label{eq:add_noise}
\end{equation}
where $t \in \{ 1 \cdots T \}$  denotes the time step of the forward diffusion process. $\mathcal{N}$ denotes the calculation following a Gaussian distribution with a mean of $\sqrt{\overline{\alpha}_t}z_0$ and a variance of $(1,\overline{\alpha}_t)\mathbf{I}$, where ${\alpha}_t := 1-{\beta}_t$ and $\overline{\alpha}_t=\prod^{t}_{s=1}{\alpha}_s$. The output $z_t$ thus follows a Gaussian distribution. To improve this accuracy, additional information, condition generated by CLIP \cite{10-clip-gal2022image} is fed into the diffusion model, as follows:
\begin{equation}
\mathcal{L}_{LDM}=\mathbb{E}_{\varepsilon(x),d,\epsilon \sim \mathcal{N}(0,1),t}[|| \epsilon - \epsilon_\theta(z_t,t,\tau_\theta(d)) ||^2_2],
    \label{eq:diffusion}
\end{equation}
where $\mathbb{E}$ denotes the mathematical expectation, $\epsilon$ represents the noise added, which follows a Gaussian distribution. $\epsilon_{\theta}$ denotes the noise predicted based on the known conditions $z_t, t, \tau_\theta(d)$, where $\tau_{\theta}$ is the image encoder from CLIP \cite{10-clip-gal2022image} and $d$ could be text prompt. Finally, the mean squared error (MSE) between the label noise and the predicted noise is calculated to obtain the loss for the LDM.

\subsection{Clothing Region of Interest Self-correction}
\subsubsection{Overview}
The framework of the proposed clothing region of interest self-correction approach (CSC) is illustrated in Fig. \ref{fig:overview}. Its inputs are clothing image, person image and the textual prompt to describe the target clothing. These inputs are first passed into the clothing diffusion model C-UNet to fuse clothing features into each layer of the person diffusion model P-UNet, outputting the generated clothing image. CSC imposes constraints on the attention map extracted during the denoising process, making the generating process place more emphasis on
the Clothing Region of Interest (C-RoI). C-ROI is the region where target clothing is synthesized onto a person after each step's sampling is completed. It shapes the attention distribution in subsequent steps, effectively addressing information loss of clothing details within C-RoI.

\noindent
\subsubsection{Energy-guided Function} 
Given the predicted noise $x_t$ of the diffusion model at the step $t$, the energy-guided function is designed as a corrective module to adjust the predicted noise $x_{t-1}$ at the next step $t-1$ during denoising, thereby guiding the diffusion model to produce better results, as follows:
\begin{equation}\label{eq:xt-1}
\begin{split}
    x_{t-1} 
    & =\mathcal{N}\big{(}
        x_{t-1};\mu_\phi(x_t,t),\Sigma_\phi(x_t,t)
        \big{)}  \\
    & =(1+\frac{1}{2}\beta_t)x_t
        +\beta_t \triangledown_{x_t}\log{p(x_t)}
        +\sqrt{\beta_t}\epsilon,
\end{split}
\end{equation}
where $\mu_\phi$ is the mean function, $\Sigma_\phi$ represents the variance function and $\beta_t \in \mathbb{R}$ is a pre-defined parameter. In \cite{yu2023freedom}, it is concluded that $\triangledown_{x_t}\log{p(o|x_t)} \propto -\triangledown_{x_t}\mathcal{E}(o,x_t)$, where $\mathcal{E}(o, x_t)$ is used to calculate the similarity between the condition $o$ and $x_t$. To adapt it to the VTON task, we replace $o$ with the warped clothing mask $M$, a binary mask where the region within C-RoI is with the value 1, while the rest region is with the value 0. Thus, it can be further concluded that $\triangledown_{x_t}\log{p(M|x_t)} \propto -\triangledown_{x_t}\mathcal{E}(M,x_t)$. By employing the energy function to correct the predicted noise 
$x_{t-1}$, the attention map can be gradually constrained within the $M$. The single-step prediction formula for $x_{t-1}$ is obtained:
\begin{equation}\label{eq:final_xt-1}
    x_{t-1}=m_t-\rho_{t}\triangledown_{x_t}\mathcal{E}(M,x_t),
\end{equation}
where $ m_t=(1+\frac{1}{2}\beta_t)x_t
        +\beta_t \triangledown_{x_t}\log{p(x_t)}
        +\sqrt{\beta_t}\epsilon $ 
represents the predicted content for the personal part, and $\rho_t$ is a scale factor. We apply this energy function to our model, using it during denoising to constrain the attention map within the mask area $M$. This ensures that during iterations, the model focuses more on the details within C-ROI.

\noindent
\subsubsection{Attention Attract Energy Function} 
For the design of $\mathcal{E}(M, x_t)$, the attention attract energy function $\mathcal{E}_{attract}$ and attention repel energy function $\mathcal{E}_{repel}$ are proposed. Firstly, as the attention map distribution tends to be scattered, the high-value points in the attention map outside $M$ should be redirected inward. In this way, $\mathcal{E}_{attract}$ aims to minimize the values of these external points:
\begin{equation}
    \mathcal{E}_{attract} = \frac{\sum_{i,j} A_{i,j}  (1 - {M}_{i,j})}{\sum_{i,j} A_{i,j}  {M}_{i,j}},
    \label{eq:E1}
\end{equation}
where $A$ denotes the attention map and $(i,j)$ is the coordinate index of points in $A$ and $M$. It is designed to concentrate attention within $M$, enhancing its generative details.

\subsubsection{Attention Repel Energy Function} Only $\mathcal{E}_{attract}$ may lead to uniform attention values within $M$, hindering comprehensive focus during image generation. To address this, $\mathcal{E}_{repel}$ is further proposed in different scenarios. For the scenario where all non-zero points in the attention map fall within $M$, indicating that the attention map has already focused on $M$, $\mathcal{E}_{repel\_inner}$ is calculated:
\begin{equation}
    \mathcal{E}_{repel\_inner} = \frac{1}{N} \sum_{i,j} \sum_{k,l} \max(0, \delta - |A_{i,j} - A_{k,l}|),
    \label{eq:E21}
\end{equation}
where $N$ represents the total number of the attention map's non-zero points within $M$; $(i,j)$, and $(k,l)$ are two sets of coordinate indexes of $A$; $\delta$ is a predefined margin value.  $\mathcal{E}_{repel\_inner}$ is used to space out attention values and alleviate the issue of uniform attention values caused by $\mathcal{E}_{attract}$.

In other scenarios, especially when most values in the attention map are 0, we propose adding a regularization term:
\begin{equation}
    \mathcal{E}_{repel\_outer} = -\sum_{i,j} A_{i,j} {M}_{i,j}.
    \label{eq:E22}
\end{equation}
$\mathcal{E}_{repel\_outer}$ is proposed to encourage more attention values to appear within $M$ and increase its attention density. In this way, the ultimate $\mathcal{E}_{repel}$ can be expressed as follows:
\begin{equation}
    \mathcal{E}_{repel} = 
    \begin{cases} 
    \mathcal{E}_{repel\_inner} & \text{if } A (1 - M) = \textbf{0}, A M \neq \textbf{0} \\
    \mathcal{E}_{repel\_outer} & \text{otherwise}
    \end{cases},
    \label{eq:E2}
\end{equation}
where \textbf{0} indicates the all-zero matrix. The purpose of $\mathcal{E}_{repel}$ is to make the attention distribution in $A$ more concentrated on C-RoI. When back-propagating the gradients for the attention map after each step, it provides a direction for aggregation. For steps where less attention is generated, the increased $\mathcal{E}_{repel}$ distance encourages more attention in the next step. Thereby it can improve the generation quality within $M$. The final optimization goal can be expressed as:
\begin{equation}
    \mathcal{E} = \mathcal{E}_{attract} + \lambda \mathcal{E}_{repel},
\end{equation}
where $\lambda$ is a weight coefficient used to balance the influence of the two aforementioned energy functions.

\begin{table*}[h!]
\small
    \setlength\tabcolsep{2pt}
    \centering

\caption{Quantitative results on the VITON-HD dataset \cite{6-choi2021viton}. The best results are reported in \textbf{bold}.}
    
    \begin{tabular}{c|cccc|cccc}
    \hline
        \multirow{2}{*}{Method}
        & \multicolumn{4}{c|}{VITON-HD (paired)} 
        & \multicolumn{4}{c}{VITON-HD (unpaired)} 
        \\ 
    \cline{2-5}\cline{6-9}
         
         &LPIPS$\downarrow$ 
         &FID$\downarrow$ &KID$\downarrow$ &VTID$\downarrow$
         &LPIPS$\downarrow$  
         &FID$\downarrow$ &KID$\downarrow$ &VTID$\downarrow$
         \\
     \hline
         HR-VITON   &0.097&12.30&0.89&16.0392 
                    &0.132&14.83&1.34&16.5459
         \\
         StableVITON    &0.084&9.13&1.20&14.2902 
                        &0.124&11.49&1.52&14.8119
         \\
         GP-VTON    &0.083
                    &9.17&0.93&15.8451
                    &0.117
                    &12.21&1.23&16.3454
        \\
         LaDI-VTON  &0.091
                    &9.31&1.53& 14.7235
                    &0.151
                    &11.93&1.80& 15.2337
        \\
         OOTDiffusion  &0.071
                    &8.81&0.82
                    &13.0472
                    &0.098
                    &9.45&1.16
                    &13.5374
        \\
    \hline
        LaDI-VTON + CSC  
                    &0.087
                    &9.16&1.41& 14.3005
                    &0.147
                    &11.62&1.71& 14.5051
        \\
        OOTDiffusion + CSC  &\textbf{0.070}
                            &\textbf{8.68}&\textbf{0.73}
                            &\textbf{12.3194} 
                            &\textbf{0.086}
                            &\textbf{9.38}&\textbf{1.03}
                            &\textbf{12.8726} 
        \\

    \hline
    \end{tabular}

    \label{tab:viton-hd}
\end{table*}

\section{Experiments}
\label{sec:experiments}
\subsection{Experimental Setup}
\label{}

\noindent
\textbf{Datasets.} 
Our experiments are conducted on two key datasets: VITON-HD \cite{6-choi2021viton} and DressCode \cite{33-morelli2022dress}. VITON-HD comprises 13,678 high-resolution images (1,024$\times$768) of individuals paired with upper-body attire, ideal for try-on simulations, with 2,032 images for model evaluation. 
DressCode offers a more diverse backdrop, with 27,261 images categorized into upper body (15,363), lower body (8,951), and dress try-ons (2,947), all with a resolution of 1,024$\times$768. 
For a fair comparison, 1,800 images from each DressCode category were selected for model assessment.



\begin{table*}[h!]
    \scriptsize
    \setlength\tabcolsep{0.5pt}   
    \centering

\caption{Quantitative results on the paired data of DressCode dataset \cite{33-morelli2022dress}. The best results are reported in \textbf{bold}.}

    \scalebox{0.93}{
        \begin{tabular}{c|cccc|cccc|cccc}
    \hline
        \multirow{2}{*}{Method}
        & \multicolumn{4}{c|}{Upper-body (paired)} 
        & \multicolumn{4}{c|}{Lower-body (paired)} 
        & \multicolumn{4}{c}{Dresses (paired)} 
        \\ 
    \cline{2-5}\cline{6-9}\cline{10-13}
         &LPIPS$\downarrow$
         &FID$\downarrow$ &KID$\downarrow$ &VTID$\downarrow$
         &LPIPS$\downarrow$
         &FID$\downarrow$ &KID$\downarrow$ &VTID$\downarrow$
         &LPIPS$\downarrow$
         &FID$\downarrow$ &KID$\downarrow$ &VTID$\downarrow$
         \\
     \hline
         GP-VTON    &0.090
                    &12.20  &1.22   &15.7414
                    &0.097
                    &16.65  &2.86   &15.8753
                    &0.101
                    &12.65  &1.84   &16.1876
        \\
         LaDI-VTON  &0.096
                    &12.30  &1.30   &15.5670
                    &0.105
                    &13.38  &1.98   &15.6992
                    &0.108
                    &13.12  &1.85   &16.0163
        \\
         OOTDiffusion  &\underline{0.079}
                    &\underline{11.03}&\underline{0.29}
                    &\underline{13.4396}
                    &\underline{0.084}
                    &\underline{9.72}&\underline{0.64} 
                    &\underline{14.8723}
                    &\underline{0.088}
                    &\underline{10.65}&\underline{0.54}
                    &\underline{13.1704}
        \\
    \hline
        LaDI-VTON + CSC  
                    &0.093
                    &12.18  &1.27   &15.1989
                    &0.103
                    &13.25  &1.70   &15.2848
                    &0.108
                    &11.93  &1.73   &15.8102
        \\
        OOTDiffusion + CSC  &\textbf{0.076}
                            &\textbf{10.11}&\textbf{0.11}
                            &\textbf{12.0213}
                            &\textbf{0.082}
                            &\textbf{8.25}&\textbf{0.29} 
                            &\textbf{12.1505}
                            &\textbf{0.085}
                            &\textbf{9.81}&\textbf{0.12}
                            &\textbf{12.4758}
        \\

    \hline
    \end{tabular}

    }

    \label{tab:dc-paired}
\end{table*}

\noindent
\textbf{Baselines.} 
This work is based on OOTDiffusion \cite{xu2024ootdiffusion} and we also compare it with recent VTON models, including StableVITON \cite{24-kim2024stableviton}, LaDI-VTON \cite{24-kim2024stableviton}, GP-VTON \cite{52-xie2023gp}, and HR-VITON \cite{27-lee2022high}.  Furthermore, since StableVITON \cite{24-kim2024stableviton} and HR-VITON \cite{27-lee2022high} are designed only for the VITON-HD \cite{6-choi2021viton} dataset, we have compared our method with OOTDiffusion \cite{xu2024ootdiffusion}, LaDI-VTON \cite{32-morelli2023ladi}, and GP-VTON \cite{52-xie2023gp} when using evaluation metrics for the DressCode \cite{33-morelli2022dress} dataset.

\begin{table*}[h!]
    \setlength\tabcolsep{0.5pt}   
    \centering
    \scriptsize

\caption{Quantitative results on the unpaired data of DressCode dataset \cite{33-morelli2022dress}. Best results are reported in \textbf{bold}.}
    \label{tab:dc-unpaired}
    
    \scalebox{0.92}{
        \begin{tabular}{c|cccc|cccc|cccc}
    \hline
        \multirow{2}{*}{Method}
        & \multicolumn{4}{c|}{Upper-body (unpaired)} 
        & \multicolumn{4}{c|}{Lower-body (unpaired)} 
        & \multicolumn{4}{c}{Dresses (unpaired)} 
        \\ 
    \cline{2-5}\cline{6-9}\cline{10-13}
         &LPIPS$\downarrow$
         &FID$\downarrow$ &KID$\downarrow$ &VTID$\downarrow$
         &LPIPS$\downarrow$
         &FID$\downarrow$ &KID$\downarrow$ &VTID$\downarrow$
         &LPIPS$\downarrow$
         &FID$\downarrow$ &KID$\downarrow$ &VTID$\downarrow$
         \\
     \hline
         GP-VTON    &0.126
                    &14.37  &1.81   &15.8766
                    &0.132
                    &17.83  &2.97   &16.0784
                    &0.137
                    &14.72  &2.14   &16.4766
        \\
         LaDI-VTON  &0.160
                    &13.59  &1.61   &15.6954
                    &0.171
                    &15.17  &2.30   &15.9106
                    &0.175
                    &15.32  &2.24   &16.2098
        \\
         OOTDiffusion  &0.106
                    &12.14&0.72
                    &13.5638
                    &0.112
                    &10.67&0.91 
                    &13.7632
                    &0.118
                    &11.85&0.80
                    &14.1114
        \\
    \hline
        LaDI-VTON + CSC  
                    &0.158
                    &13.27  &1.58   &15.4270
                    &0.169
                    &15.04  &2.11   &15.5116
                    &0.168
                    &15.19  &2.12   &16.0117
        \\
        OOTDiffusion + CSC  &\textbf{0.095}
                            &\textbf{10.84}&\textbf{0.41}
                            &\textbf{12.1554}
                            &\textbf{0.097}
                            &\textbf{10.03}&\textbf{0.76} 
                            &\textbf{12.3645}
                            &\textbf{0.102}
                            &\textbf{10.32}&\textbf{0.79}
                            &\textbf{12.7524}
        \\

    \hline
    \end{tabular}

    }

\end{table*}

\noindent
\textbf{Implementation Detail.} 
AdamW \cite{29-loshchilov2017fixing} is used as the optimizer with the learning rate as 5e-5. The guidance scale for classifier-free \cite{20-ho2022classifier} guidance is set to 2.0 by default, and $\rho_t$=0.2, $\delta$=0.02, $\lambda$=0.01. The model is trained on the VITON-HD \cite{6-choi2021viton} dataset at a resolution of $512\times 384$ with a batch size of 64 and on the DressCode \cite{33-morelli2022dress} dataset at a resolution of $1024\times 768$ with a batch size of 16 for 36,000 iterations. In inference, we set the sampling steps to 20, using the UniPC sampler \cite{59-zhao2024unipc}. The GPU requirement for both inference and training is one NVIDIA A100 GPU.

\begin{figure}[h!]
    \centering
    \includegraphics[width=0.9\linewidth]{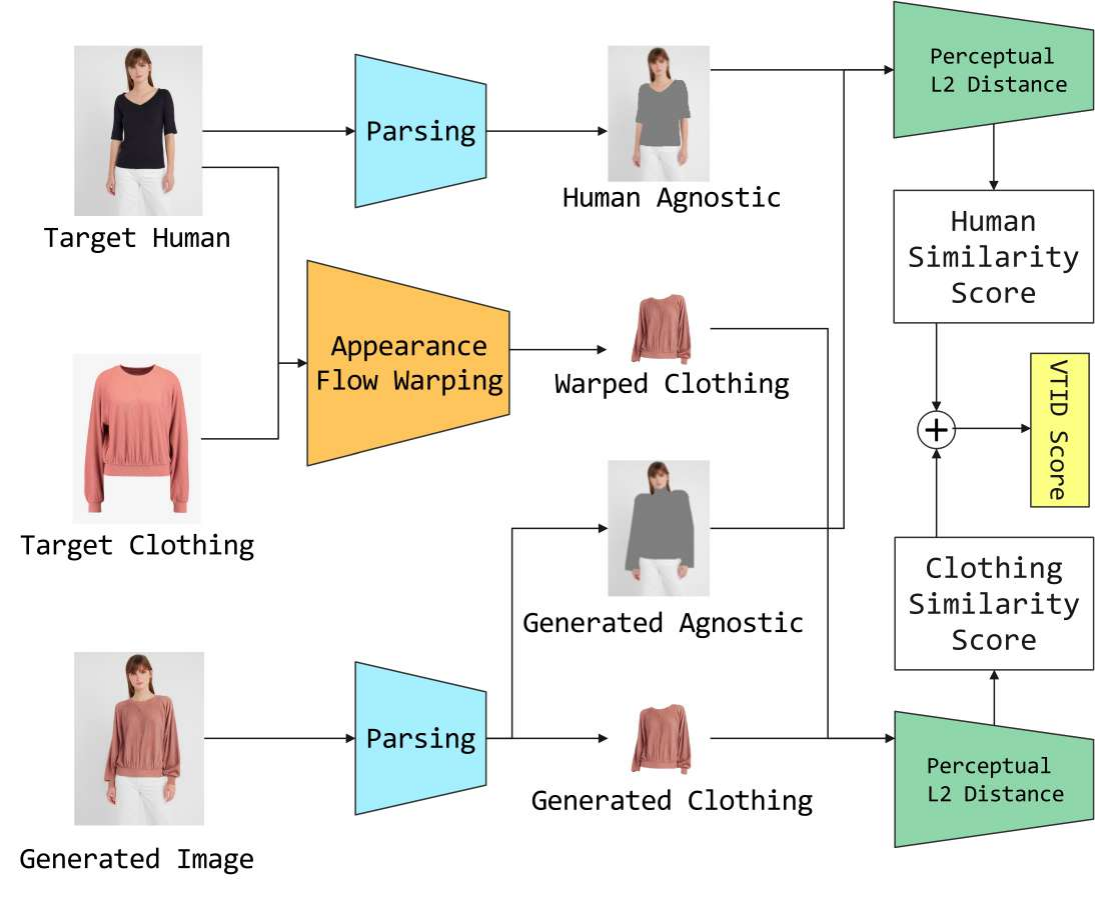}
    \caption{Illustration of the new evaluation metric VTID. }
    \label{fig:vid}
\end{figure}

\noindent
\textbf{Metrics.} 
Traditional metrics focus on both paired and unpaired data synthesis. For paired data, metrics such as LPIPS \cite{58-zhang2018unreasonable}, FID \cite{19-heusel2017gans} and KID \cite{2-binkowski2018demystifying} are standard for assessing synthesis quality. However, in the context of unpaired data in VTON applications, where outputs may significantly vary from the original inputs, the conventional FID metric's reliability in indicating synthesis quality is compromised.

\begin{figure*}[!h]
    \centering
    \includegraphics[width=1\linewidth]{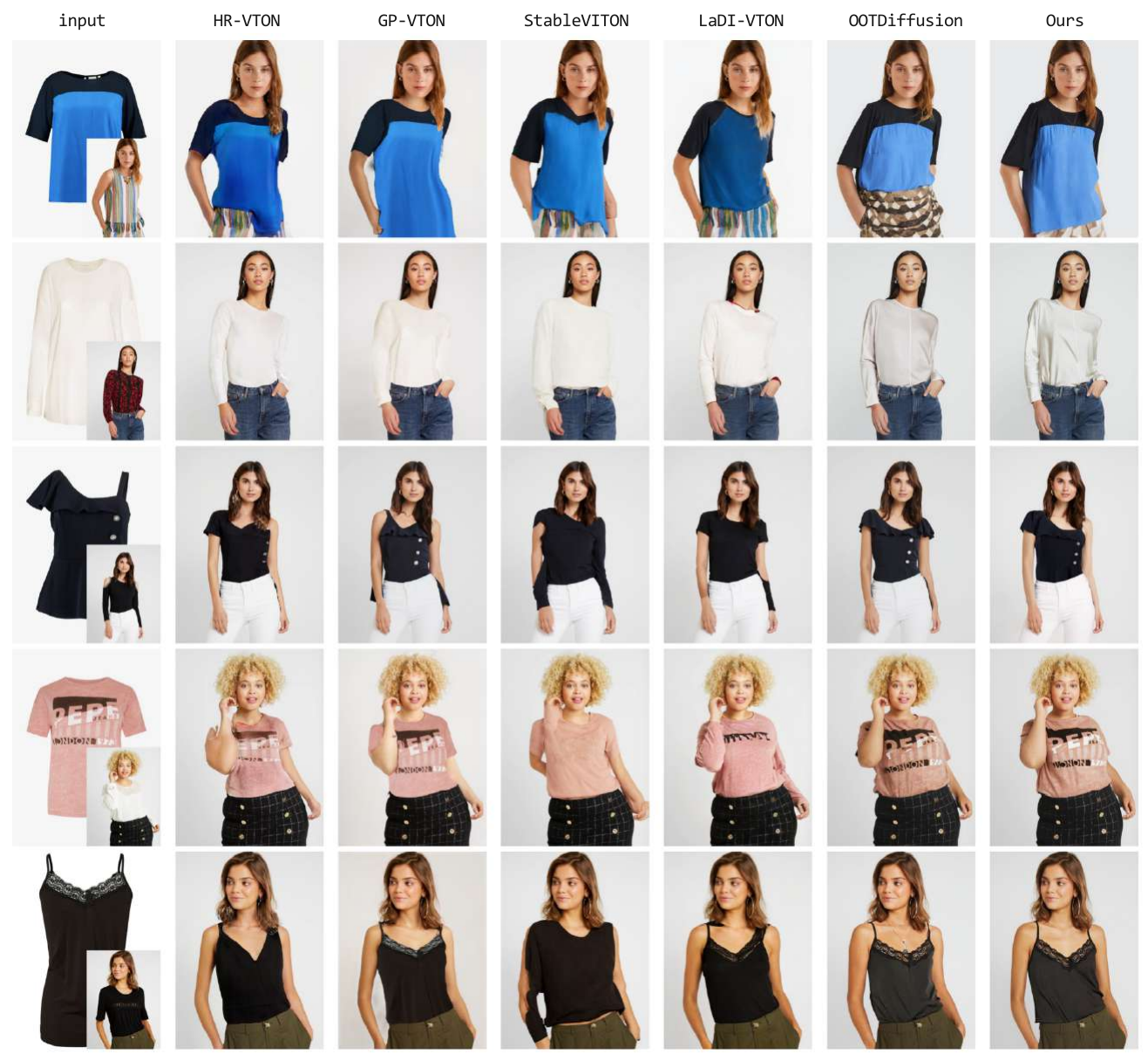}
    \caption{Qualitative comparison on the VITON-HD dataset. Please zoom in for more details.}
    \label{fig:test_qualitative}
\end{figure*}

For unpaired data, our evaluation focuses on preserving the original human pose and clothing appearance during virtual try-ons. The VTID metric, illustrated in Fig.~\ref{fig:vid}, assesses both paired and unpaired synthesis. Using human parsing~\cite{12-ge2021parser} and appearance flow warping~\cite{16-han2018viton}, we generate four representations: human agnostic, warped clothing, generated agnostic, and generated clothing. We then compute perceptual L2 distances~\cite{vggloss-johnson2016perceptual} between corresponding pairs to obtain human and clothing similarity scores, measuring pose and appearance consistency. The final VTID score combines these two metrics.



\subsection{Quality of Generated Clothing Images}
\label{sec:quality}

\noindent
\textbf{Quantitative Resuls.}
We divide the VITON-HD and DressCode into paired and unpaired subsets, applying our model to each and evaluating using LPIPS, FID, KID and VTID. Our model, combined with OOTDiffusion and CSC, achieves the best performance in most metrics for both paired and unpaired VITON-HD, as shown in Table \ref{tab:viton-hd}.
Incorporating the CSC method into LaDIVTON, we note metric improvements over the baseline models. For the DressCode dataset, separated into upper body, lower body, and dresses, the results are presented in Tables \ref{tab:dc-paired} and \ref{tab:dc-unpaired}. Notably, HR-VITON and StableVITON are not tested on this dataset. The upper body try-ons in Table \ref{tab:dc-paired} show less favorable results, likely due to the complexity of occlusion scenarios. In contrast, the lower body category, with its limited variations and occlusions, achieves the best outcomes.

As depicted in Table \ref{tab:dc-unpaired}, the assessment of unpaired try-on outcomes lags behind those of paired data, underscoring the inherent VTON domain challenge where pre- and post-try-on states inherently differ. Notably, the integration of OOTDiffusion with CSC in the unpaired dataset results in noticeable improvements over the baseline and other models. The proposed VTID metric unifies the measurements of both paired and unpaired evaluations, showing its reliability. 

\noindent
\textbf{Qualitative Results.}
The comparison of the generated results from multiple models can be seen in Fig. \ref{fig:test_qualitative}. Due to space limitations, the Figure only displays the comparative results of several existing models, with the detailed content available in the \textit{supplementary material}. The first row in Fig. \ref{fig:test_qualitative}, shows the person wearing a dress, with the target clothing being a top. All models compared with ours struggled to accurately generate the boundaries, resulting in the clothing either exceeding the boundaries or not fitting the upper body properly. 
In the second row, the texture of the target clothing is lost in non-diffusion models, leading to only the color being retained. The third and fourth row demonstrates the ability to preserve structure and complex patterns. The fifth row involves testing the generation of details and lace edges.
In summary, the figure demonstrates that our proposed CSC method, by aggregating the attention map, enables diffusion-based VTON models to generate better results within the clothing region of interest, including improved boundary constraints, texture quality, pattern retention, and reduced loss of positional information.

\begin{figure*}[!h]
    \centering
    \includegraphics[width=0.9\textwidth]{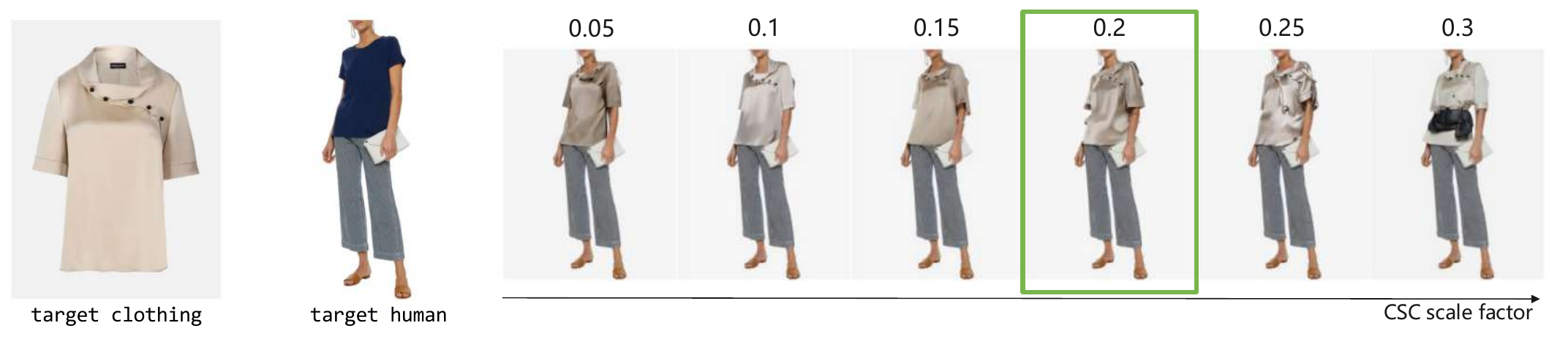}
    \caption{Illustration of the synthetic results with different scale factor values.  Please zoom in for more details.}
    \label{fig:ablation-factor}
\end{figure*}
\begin{figure*}[!h]
    \centering
    \includegraphics[width=0.9\textwidth]{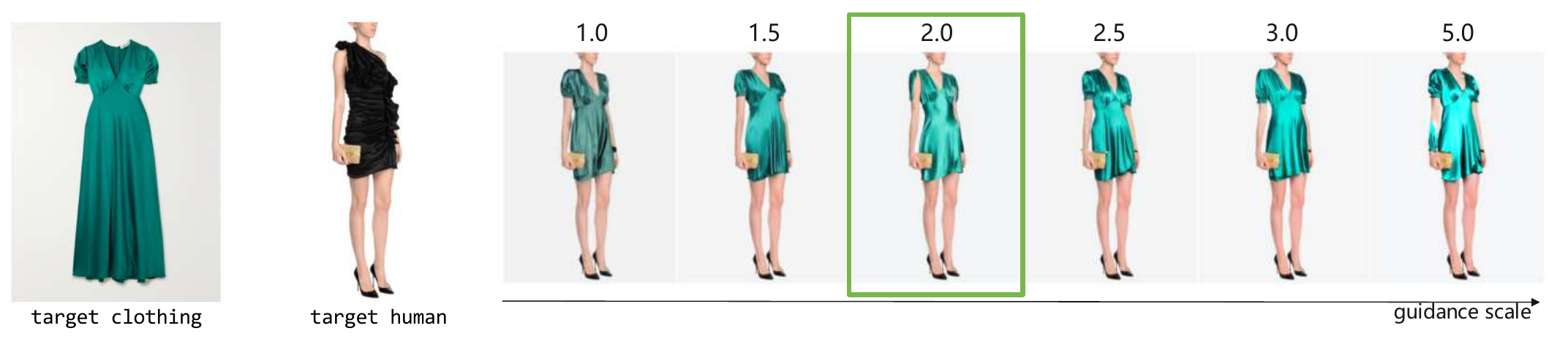}
    \caption{Illustration of the synthetic results with different guidance scale values.  Please zoom in for more details.}
    \label{fig:ablation-scale}
\end{figure*}


\subsection{Effectiveness for Downstream Task}
\label{sec:application}
The proposed model is also applied to alleviate the data limitation issue of the downstream task, such as CC-ReID, which identifies individuals across various cameras regardless of clothing changes. By enriching the variety of clothing types for individuals \cite{extra2024-reid-aug}, we expand three major datasets (LTCC \cite{ltcc-qian2020long}, PRCC \cite{prcc-yang2019person}, Vc-Clothes \cite{vcc-wan2020person}) and retrain the previous SOTA model SCNet \cite{ccreid-scnetguo2023}, surpassing the original accuracies of SCNet in all metrics.

\begin{figure}[htb]
    \centering
    \includegraphics[width=0.9\linewidth]{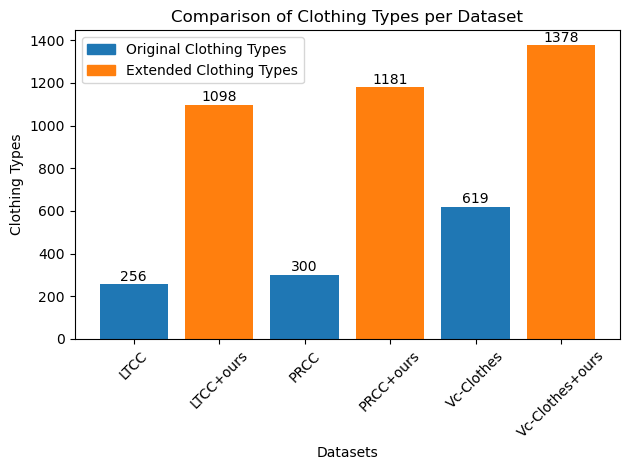}
    \caption{Illustration of the types of clothing included in the LTCC, PRCC, Vc-Clothes datasets before and after expansion.}
    \label{fig:clothing_types}
\end{figure}

\begin{table}[h!]
    \centering
    \setlength\tabcolsep{6.5pt}   
    
    \caption{Comparison performance of multiple ReID models on three datasets, ``+ours'' represents the results of the model optimized with the training data generated by our method.}

    \scalebox{1.04}{
        \begin{tabular}{c|cc|cc|cc}
    \hline
        \multirow{2}{*}{Method}
        & \multicolumn{2}{c|}{LTCC} 
        & \multicolumn{2}{c|}{PRCC} 
        & \multicolumn{2}{c}{Vc-Clothes} 
        \\ 
    \cline{2-3}\cline{4-5}\cline{6-7}
        &Rank-1$\uparrow$ &mAP$\uparrow$ 
        &Rank-1$\uparrow$ &mAP$\uparrow$ 
        &Rank-1$\uparrow$ &mAP$\uparrow$ 
         \\
     \hline
         HACNN\tiny{CVPR18}    
                    &21.6  &9.3   
                    &21.8  &23.2
                    &49.6  &50.1
        \\
         PCB\tiny{ECCV18}
                    &23.5  &10.0
                    &41.8  &38.7
                    &62.0  &62.2
        \\
         IANet\tiny{CVPR19}
                    &25.0  &12.6
                    &46.3  &45.9
                    &-  &-
        \\
         ISP\tiny{ECCV20} 
                    &27.8  &11.9
                    &36.6  &-
                    &72.0  &72.1
        \\
    \hline
         FSAM\tiny{CVPR21}
                    &38.5  &16.2
                    &-  &-
                    &78.6  &78.9
        \\
         CAL\tiny{CVPR22}
                    &40.1  &18.0
                    &55.2  &55.8
                    &81.4  &81.7
        \\
         GI-ReID\tiny{CVPR22} 
                    &23.7  &10.4
                    &-  &-
                    &64.5  &57.8
        \\
         AIM\tiny{CVPR23} 
                    &40.6  &19.1
                    &57.9  &58.3
                    &-  &-
        \\
         CCFA\tiny{CVPR23} 
                    &45.3  &22.1
                    &61.2  &58.4
                    &-  &-
        \\
        SCNet\tiny{ACM2023}  
                            &47.5  &25.5
                            &61.3  &59.9
                            &90.1  &84.4
        \\
    \hline
        SCNet + ours       
                            &\textbf{48.4}  &\textbf{26.5}
                            &\textbf{62.7}  &\textbf{61.2}
                            &\textbf{90.8}  &\textbf{85.8}
        \\
    \hline
    \end{tabular}

    }

    \label{tab:reid}
\end{table}

Based on the original pipeline, we designed an optimization method tailored for the multi-clothing ReID task by ensuring that each identity wears different clothing in each frame. This approach, which greatly diversifies clothing while still corresponding to a single identity, causes the ReID model to gradually treat clothing as an irrelevant variable. The results of this method can be seen in Table \ref{tab:reid}.

Our method applies augmentation exclusively to the training sets of the three datasets to ensure the validity of the results. As shown in Fig. \ref{fig:clothing_types} and Table \ref{tab:reid}, there is a large increase in the number of clothing types for each dataset's training set, leading to corresponding improvements in the Rank-1 and mAP scores. These results demonstrate the effectiveness and feasibility of our approach.

\begin{figure}[!t]
    \centering
    \includegraphics[width=1\linewidth]{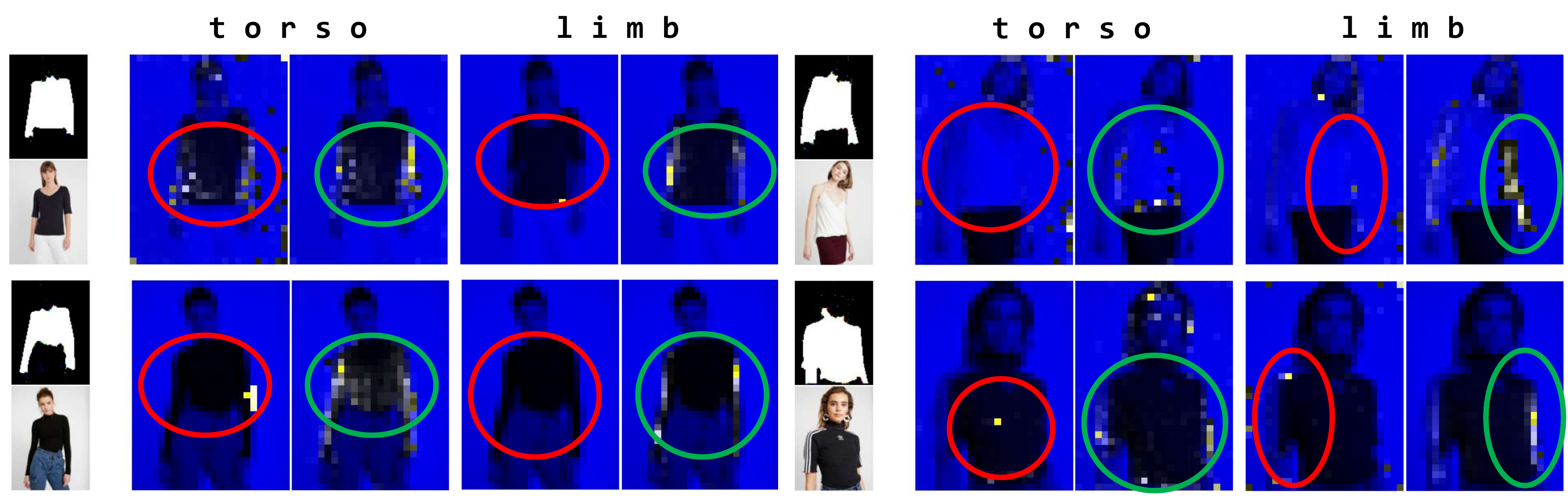}
    \caption{Ablation study on the distribution of attention across different human parts without/with CSC. The two rows of images show the distribution of attention maps on the input for randomly selected steps. The image within the red circle is ``without CSC'', and the image within the green circle is ``with CSC''.}
    \label{fig:attn}
\end{figure}

\subsection{Ablation Study}
\label{sec:ablation}

The ablation studies of the proposed method are conducted on three aspects: setting different scale factors, selecting different attention layers and comparing different guidance scales to observe the quality of experimental results. 


\begin{table}[h!]
    \setlength{\tabcolsep}{16pt}

    \caption{Ablation study of different scale factors on the VITON-HD dataset with guidance scale value=2.0.The best and second best results are reported in \textbf{bold} and \underline{underline},  respectively.}

    \scalebox{1.005}{
        \begin{tabular}{c|cccc}
    \hline
         CSC scale factor
         &LPIPS$\downarrow$                   
         &FID$\downarrow$ &KID$\downarrow$ &VTID$\downarrow$  \\
    \hline
         -      &\underline{0.0710}&\underline{8.81}
                    &0.82&13.0472 \\
         0.05 &0.0740&8.92&0.81&13.1042 \\
         0.1    &0.0738&8.90&0.80&12.8007 \\
         0.15    &0.0737&8.89
                    &\underline{0.79}&\underline{12.4066} \\
         0.2    &\textbf{0.0704}&\textbf{8.68}
                    &\textbf{0.73}&\textbf{12.3194} \\
         0.25    &0.0723
                    &8.85&0.82&13.0062 \\
         0.3    &0.0749&8.91&0.98&13.2611 \\

    \hline
    \end{tabular}

    }

    \label{tab:CSC-scale}
\end{table}

\noindent
\textbf{Different scale factors.}
The first ablations study is to explore different scale factors. In Table \ref{tab:CSC-scale}, the first row indicates that the CSC scale factor is not used, which is equivalent to a factor of 0. This means that no adjustment is made to $x_{t-1}$ using CSC, which is equivalent to the effect of the baseline. Then we set the CSC scale factor to 0.05, 0.1, 0.15, 0.2, 0.25, and 0.3, comparing each other with the guidance scale of 2.0. Fig. \ref{fig:ablation-factor} shows the results synthesized with the CSC scale factors set to 0.05, 0.1, 0.15, 0.2, 0.25, and 0.3, respectively. As the factor increases, the quality of the image synthesis first improves and then declines. At 0.2, the balance between lighting and texture results in a better synthesis effect. Additionally, we visualized the attention maps before and after applying CSC. As shown in Fig. \ref{fig:attn}, CSC effectively encourages the generation and aggregation of more attention to regions of interest.

\begin{table}[!h] 
    \setlength{\tabcolsep}{21pt}
    \centering

        \caption{Ablation study of using different attention layers. The table presents the evaluation results for calculating CSC using different attention layers. The best and second best results are reported in \textbf{bold} and \underline{underline}, respectively.}

    \begin{tabular}{c|ccc}
    \hline
        Attention layer& FID$\downarrow$ & KID$\downarrow$ 
                        & VTID$\downarrow$ 
                         \\
    \hline    
        Full Use & \textbf{8.65} & \textbf{0.69} & \textbf{12.3176}  \\
        Only Downsampling & 8.71 & 0.77 & 13.4750  \\
        Only Upsampling & 8.72 & 0.75 & 13.4807  \\
        Paired 12,288 (128$\times$96) & 9.53 & 1.08 & 15.0885  \\
        Paired 3,076 (64$\times$48) & 9.12 & 0.81 & 14.8701  \\
        Paired 768 (32$\times$24) & \underline{8.68} & \underline{0.73} 
                            & \underline{12.3194} 
                            \\
        Paired 192 (16$\times$12) & 9.15 & 0.90 & 14.0052 
                            \\
    \hline
    \end{tabular}

    \label{tab:attn_layer}
\end{table}

\begin{table}[h!]
\setlength{\tabcolsep}{17.55pt}
    \caption{Ablation study of different guidance scale values on the VITON-HD dataset with CSC scale factor=0.2.The best and second best results are reported in \textbf{bold} and \underline{underline}, respectively.}
    \begin{tabular}{c|cccc}
    \hline
         guidance scale
         &LPIPS$\downarrow$                   
         &FID$\downarrow$ &KID$\downarrow$ &VTID$\downarrow$  \\
    \hline
         -      &0.1021&12.57
                    &0.96&15.7302 \\
         1.0 &0.0879&10.79&0.91&14.2882 \\
         1.5    &0.0750&\underline{8.71}
                &\underline{0.76}&\underline{12.4002} \\
         2.0    &\textbf{0.0704}&\textbf{8.68}
                    &\textbf{0.73}&\textbf{12.3194} \\
         2.5    &\underline{0.0745}&8.78
                    &0.77&12.4570 \\
         3.0    &0.0821
                    &8.83&0.80&12.8020 \\
         5.0    &0.0932&9.31&0.83&13.9341 \\

    \hline
    \end{tabular}

    \label{tab:guidance-scale}
\end{table}

\noindent
\textbf{Different attention layers.}
Table \ref{tab:attn_layer} displays the layer selection for CSC calculation, with the first setting using all attention maps. In the second and third settings, attention maps are utilized during only the upsampling or downsampling phases, respectively. The subsequent settings detail a targeted strategy that incorporates attention maps from both upsampling and downsampling that fulfill precise dimensional standards for effective correction. Ultimately, the ``Full Use'' option is endorsed for its ability to fully engage all potential insights within the model.

\noindent
\textbf{Different guidance scales.}
The guidance scale is used to adjust the ratio between condition-based and condition-free, balancing the degree of freedom in the diffusion model during image generation. We conduct ablation experiments on the guidance scale to verify its role in our proposed method with a fixed CSC scale factor of 0.2. In Table \ref{tab:guidance-scale}, when the guidance scale is set as 1, it indicates that the guidance scale is not used, which is equivalent to not employing classifier-free guidance. As can be seen in Fig. \ref{fig:ablation-scale}, the results reach a critical point when the value is 2.0. At this point, the retention of the clothing's style and texture reaches an equilibrium, closely resembling the ground truth.

\section{Conclusion}
\label{}
This paper presents the CSC to refine the diffusion model's generation process by focusing on key clothing features within the attention map. Our method significantly improves the VTON model's ability to self-correct and preserve these features, as evidenced by experiments on the OOTDiffusion baseline. We also introduced the metric, named VTID, and validated its reliability through comparative evaluations. By integrating our optimized VTON model into the CC-ReID task, we further elevated the SOTA performance.
\section{Acknowledgments}
This work was supported by Young Scientists Fund of the National Natural Science Foundation of China (Grant No.62302328), Jiangsu Province Foundation for Young Scientists (Grant No. BK20230482), Suzhou Key Laboratory Open Project (Grant No. 25SZZD07)and Jiangsu Manufacturing Strong Province Construction Special Fund Project (Grant Name: Research and Development and Industrialization of Intelligent Service Robots Integrating Large Model and Multimodal Technology).

\bibliographystyle{splncs04}
\bibliography{mybib}
%





\end{document}